# Turing: Then, Now and Still Key

Kieran Greer, Distributed Computing Systems, Belfast, UK.
http://distributedcomputingsystems.co.uk

**Abstract:** This paper[1] looks at Turing's postulations about Artificial Intelligence in his paper 'Computing Machinery and Intelligence', published in 1950. It notes how accurate they were and how relevant they still are today. This paper notes the arguments and mechanisms that he suggested and tries to expand on them further. The paper however is mostly about describing the essential ingredients for building an intelligent model and the problems related with that. The discussion includes recent work by the author himself, who adds his own thoughts on the matter that come from a purely technical investigation into the problem. These are personal and quite speculative, but provide an interesting insight into the mechanisms that might be used for building an intelligent system.

## 1  Introduction

The idea of artificial intelligence has been around since the 1950's at least. Alan Turing and others have been attributed as the founders of the science and Turing as the father of AI, but a definition of what it represents is still not clear. Turing defined it through the imitation game, where a human and a machine are asked the same set of questions. If the interrogator cannot tell which is the human and which is the machine, then the machine is considered to be intelligent. This is really the ultimate test for an intelligent system, where it acts almost exactly as a human would. Most people would recognise that if a machine can perform more simplistic, but still intelligent acts, then it is considered to have intelligence.

---

There are now many different variations and definitions of what a single intelligent act might be, which is probably why a concise definition is so difficult. It probably requires however that the machine can do something by itself, without being told exactly how to do it first. This paper considers Turing's 'Computing Machinery and Intelligence' paper [22], which is one of the first to write about artificial intelligence. It looks at the postulations made in that and describes how relevant they still are today. While the problem of what artificial intelligence is and what it requires has now been defined much more formally, we are still not much further on at achieving it in a real sense. Many smaller advances have been made however and are covered in many texts about the topic.

Turing stated that a computer using a ticker tape as the information source, if programmed properly, would be able to solve any problem. This was described as the Turing machine, which is able simply to manipulate symbols on a strip of tape according to a set of rules. This can be used to simulate the logic of any computer algorithm, because a computer process is eventually broken down into simplistic on/off commands. This could be represented by a hole or not in the tape, for example. Turing meant that if a problem can be described in terms of a computer algorithm, then the computer can perform that task. The computer does not have any difficulty performing the most complex of calculations, but it does not know how to write the algorithm telling it how to do them in the first place. That requires intelligence. Neurons in an animal brain are also by nature very simple components. They switch on, or fire, when the input that they receive exceeds a certain threshold. This does not appear to require any inherent intelligence either and looks more like a purely mechanical process. The problem is that it is still not known how to use this sort of entity to realise a truly intelligent system.

The brain is made up of billions of these neurons [24]. If a single neuron has zero intelligence, then some thousands of them together also have zero intelligence; but we know that intelligence is made up of the collective activity of a large number of neurons, all firing together. One key factor is the granularity of the patterns that they form. The patterns are made up of so many neurons that a slight change in an input stimulus could lead to noticeable changes in the patterns that are produced and therefore in the output signals that they create. Different patterns can then be interpreted by the brain as something different. The brain creates electrical signals, causing changes in some internal state that might be felt. If the brain can recognise these different feelings or reactions, to small pattern changes, they could also be remembered and linked, to form different memories. Is this the essence of what intelligence is? Is it the ability to recognise these differences in a coherent and consistent way? The fact that a stimulus is involved is probably not a surprise. The electrical signals would excite and the brain would probably register this in some way. What might be surprising is how important a role it plays, but this theory appears to be a part of the mainstream research interest. The paper [23] discusses it and describes the difficulties with measuring stimulus-driven responses, or modelling the neural circuits produced by them. If we assume that individual neurons are not intelligent by themselves, then we have the following problems and requirements for realising an intelligent machine:

1. The formation of the brain structure could be driven mainly by stimuli – sight, sound, taste, touch, smell, for example. The brain tries to remember and repeat the ones that it finds favourable, or unfavourable.
2. If the neurons have no intelligence, then at least one other layer that can interpret what the firing neurons signal is required. A layer called the neocortex [8] is already known to perform more intelligent processing. The neocortex is a thin layer above the main brain body. It contains most of the intelligent components, including memory and is the driving force behind intelligent activity.
3. If external stimuli control everything about how the brain forms then environmental factors are too critical and so this more intelligent and independent layer is very important.
4. With a largely unintelligent main brain body, notions about the sub-conscious are also possible.

The following questions also arise:
1. Is the brain formation driven by the stimulus or by the more intelligent layer?
2. Does the more intelligent layer simply interpret different signals, or can it have an influence over their creation and activation?
3. Is there a process of reading and re-organising existing patterns, which would indicate a controlling and therefore intelligent process? Does the main brain body form certain patterns that are read by other layer(s) that then form other patterns, etc., until the refinement forms into comprehensible intelligence?

4. The problem is then the act of thinking itself. Without any particular external stimulus, we still think about things. So the stimulus can also be generated internally. Can a stimulus result be learnt and remembered? What sort of stimulus would allow a person to learn mathematics, for example?
5. Memory plays a key part in retrieving already stored patterns, but how do we think over them and change them in an intelligent way? Memory must also accommodate the processes to do that.

The rest of this paper is organised as follows: Section 2 gives an introductory overview of the human brain neuron and its artificial equivalent. Section 3 lists certain requirements for an intelligent system. Section 4 then lists general mechanisms or processes for finding a solution to these. Section 5 describes some work by the author that is related to an intelligent system, while section 6 gives some conclusions on what has been written about. Turing's postulations are noted in places throughout the paper, but the paper is more of a summary on general conditions for artificial intelligence. The conclusions section however lists the postulations in detail and notes how important they still are.

## 2   Simplified Model of the Human Brain

This section does not try to describe the brain structure completely, or in detail, but instead will attempt to show the similarities between the biological and the simulated neuronal processing unit. The Introduction has already described how intelligence is realised through the collective activities of these neurons, firing in response to sensory input. The paper [23] notes that while responses to sensory input can account for something such as vision, the majority of brain activity is generated internally and is silent. The brain therefore also needs to be able to produce its own activity, in the absence of external input. We therefore need to be able to think constructively.

The most obvious way to try and realise intelligence in a machine is by copying the most intelligent thing that we know of, which is the human brain. If we can figure out how that works and reproduce it in a machine, then we will have artificial intelligence. There has been quite a lot of work carried out looking at natural processes and then trying to copy them in machines. They can be called 'bio-inspired' and the idea is that if nature has worked out how to do something well, then it probably cannot be beaten and so we should try to copy it in some way. Often however the goal is not to copy it exactly, but to try to understand the underlying process that is happening and then try to implement that in a machine in some way ([5], section 1.2, for example). Computer chess, however, is a classic example of how different the result can be. Computers can now play the game of chess as well as any human and the computer programs were built around the fundamental principles of how we play the game. The machine evaluates each position and searches logically through sequences of moves to find the best possible reply in any position. However, while humans evaluate tens of positions before making a move, the best computer program can evaluate thousands or more of positions to achieve the same result. The assumption must be that the computer does not understand the problem in the same way that the human player does and therefore needs to evaluate many more positions to come to the same conclusion.

The basic computational unit in the brain is the neuron. A neuron has inputs called dendrites, a cell body and an output called an axon. An animal or human neuron can be modelled as in Figure 1. The dendrites send incoming signals into the cell body of the neuron that can be of varying strength. A stronger input signal will trigger the neuron's own output signal more. When the accumulative value of these signals exceeds a certain threshold value, the neuron fires by sending out its own signal through the axon. This output signal can then act as an input signal to other neurons, and so on.

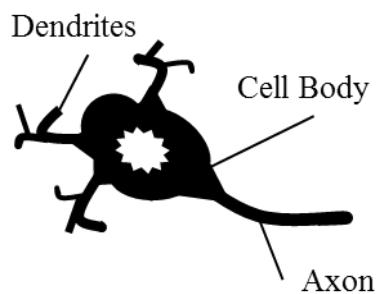

**Figure 1. Example of a human neuron.**

The signals are electrical and sent by neurotransmitters to other brain areas. They are created by chemical reactions that cause a diffusion of positive and negative ions. The positively charged ions create the electrical signal and a firing spiking event. This spiking event is also called depolarization, and is followed by a refractory period, during which the neuron is unable to fire. This could be important, because after a cell fires, it cannot then fire again, through feedback for example, before a period of time has elapsed. This could help to prevent cycling, for example. As written about in [21]

'Although the models which have been proposed to explain the structure of the brain and the nervous system of some animals are different in many respects, there is a general consensus that the essence of the operation of neural ensembles is "control through communication". Animal nervous systems are composed of thousands or millions of interconnected cells. Each one of them is a very complex arrangement which deals with incoming signals in many different ways. However, neurons are rather slow when compared to electronic logic gates. These can achieve switching times of a few nanoseconds, whereas neurons need several milliseconds to react to a stimulus. Nevertheless, the brain is capable of solving problems that no digital computer can yet efficiently deal with. Massive and herarchical networking of the brain seems to be the fundamental precondition for the emergence of consciousness and complex behaviour.'

Neural networks are the technology that most closely map to the human brain. They are the original attempt to build a machine that behaves in the same way. The inspiration for neural networks comes from the fact that although current computers are capable of vast calculations at speeds far in excess of the human brain, there are still some operations (such as speech, vision and common-sense reasoning) that current AI systems have trouble with. It is thought that the structure of the human brain may be better suited to these tasks than a traditional computing system and a neural network is an attempt to take advantage of this structure. There are many texts on neural networks, for example, [7] or [21]. In [7] the definition of a neural network is given as:

'A neural network is an interconnected assembly of simple processing elements, units or nodes, whose functionality is loosely based on the animal neuron. The processing ability of the network is stored in the interunit connection strengths, or weights, obtained by a process of adaption to, or learning from, a set of training patterns.'

Figure 2 shows a model that has been used with artificial neural networks, with the areas related to a real neuron in brackets. This shows a number of inputs ($X_1$ to $X_n$) that are weighted ($w_1$ to $w_n$) and summed, before being passed through a threshold function. If the total sum is larger than the threshold, the neuron will 'fire', or send an output signal. If it is less, then the neuron will not fire. This is an example of a perceptron, which is one of the earliest artificial neuronal models, based on McCulloch and Pitts [16]. It is described here to show how similar in nature it is to the real neuron. Neural network models, also known as connectionist models, consist of a large number of these simple processing elements, all operating in parallel. A large number of weighted connections between the elements then encode the knowledge of the network. The problem to be solved is also distributed across all of the processing elements, where it is broken down into much simpler functions. A learning algorithm is then used to adjust the weight values until the neural network has correctly learned the global function.

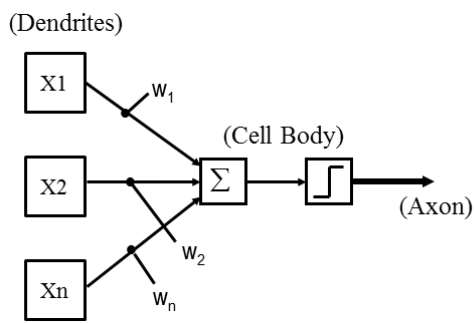

**Figure 2. Example of an artificial neuron.**

Neural networks can be used in different ways. They are very good for pattern recognition and therefore can be used simply as statistical classifiers in engineering fields. They can perform certain classification tasks better than other alternatives. As part of a model for intelligence however, they also map closely to the human brain, although the main statistical process of weight reinforcement is still too simplistic to model the real neuron properly [7]. A global function is therefore created from the combined results of simpler functions, each representing a single processing unit. These units can also be placed in layers that result in more complex representations, as the input/output flows through each one. An example of a 3-layer feedforward neural network is shown in Figure 3. Each element in one layer sends its output to every element in the next layer. All inputs to any element are weighted, summed together and then passed through an activation function to produce the output for that element. The nodes in the hidden layers may represent complex features that are discovered by the learning algorithm of the network. It is generally not known beforehand exactly what the hidden layers represent and so neural network researchers tend to characterise them in terms of their statistical properties, rather than in terms of symbolic meaning.

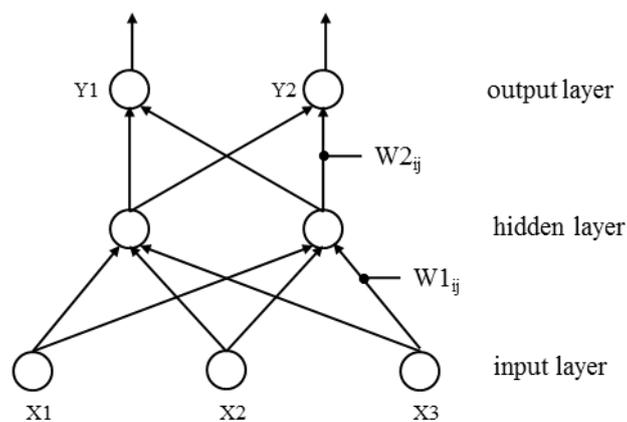

**Figure 3. Feedforward neural network.**

A neural network must be trained before it can be used. This is done by presenting it with data that it then tries to repeat, but in a general way. The learning process can also be supervised or unsupervised. In a supervised training methodology, input values are fed into the network and the resulting output values are measured. These actual output values should match a desired set of output values that are also specified as part of the training dataset. The errors, or differences, between the desired and the actual output values are then fed back through the network, to adjust the weight values at each node. Adjusting the weight values for each node will then change the output value that the node produces. This will then change the actual output values of the neural network, until they are found to be correct. After the network has been trained to recognise a training set of patterns, it is tested with a different set of test patterns. If it can also successfully classify the test set, then the network is considered to have been properly trained. The test set would be different to the training set and so they can only be correctly classified if the network has learned to generalise over the different patterns, rather than rote learning the training dataset. This generalisation has in effect allowed the network to learn the function that maps the data input values to the data output values for the type of problem being specified.

Unsupervised learning means that there is not a definite set of output values that must be matched, when in that case, the output error can be measured. Instead, the network continues to learn and adjust its values until it settles on a stable state. The network starts with some sort of hypothesis, or set of values, when neighbouring nodes then compete in their activities through mutual interactions, to best match the input data. Errors in the matching process update weight values, until a more stable state is reached. This causes the individual nodes to adapt into specific detectors of different signal patterns. Supervised learning therefore allows a neural network to recognise known patterns, while unsupervised allows it to find unknown patterns in the data. The process could be looked at as trying to minimise the error in the whole system (the neural network), or trying to realise a more stable state. When the output becomes accurate enough, the error is minimised and further corrections are not required.

While neural networks are not particularly intelligent; their highly distributed design with relatively simple individual components, makes them an enduringly attractive model for trying to mimic intelligence and several variations of the original model have since been developed. Feedforward neural networks [25] are more often associated with supervised learning, while self-organising ones [14] are more often associated with unsupervised learning. The distributed model has also been extended with other types of component, into systems such as agent-based [13] or autonomous [12] ones. With these, each individual component can be more complex. It can have its own internal reasoning engine and make decisions for itself. The overall nature of the system however is still to realise more complex behaviours or activities through distributed communication and cooperation. The reason being that the problem itself is too complex to be modelled in its entirety and so lots of simpler components are required to try to work the problem out through adaption and cooperation. So these systems already model more complex components as the individual entities and the idea of seeing the 'big' in the 'small' is also a part of nature. If a single neuron can be considered as an individual entity that produces an output, why not see a collection of neurons firing together also as an individual entity that produces an output? Then the model becomes much more complex, but still based on the same set of components.

## 3   A List of Requirements for Intelligence

This section lists a set of functionality that an intelligent system is thought to require. If you were going to build an intelligent system, it would need to include probably most of the following items. Although we know what the brain is made of physically, there are also a number of general functional requirements for what we understand intelligence to be. These are really what have been worked on over the years in AI and so the key functionality of intelligence is now well defined. Computer systems that are built can range from single complex components to a number of highly distributed and more simplistic ones. These can simply react to an input and perform some sort of statistical update, or have internal knowledge and be able to make decisions. The centralised approaches are more closely associated with knowledge-based methods, that is, methods that use existing knowledge. Distributed approaches are more closely associated with experience-based methods, that is, methods that require feedback or experience from use, to update related values. A centralised approach would include a knowledge-base or rule-based expert system [19], for example. A distributed approach would include a neural network ([7], [16], [21] or [25]) or agent-based system [13], for example. This paper deals more with the distributed options as they model the real brain more closely; however the different approaches are used to build different types of system and so cannot be compared directly in that respect. Either type has advantages and disadvantages. If you are asking a system to model a well understood problem based on certain criteria, you require a single knowledgeable system that can apply its knowledge to your problem. If you are asking a system to model a less well understood problem, you might require several distributed entities that can interact in different ways, to play out as yet unforeseen scenarios.

Relatively simple entities can be shown to exhibit more intelligent behaviour collectively, where they can use cooperation to compete with a more knowledgeable and centralised system. A centralised system can store a large amount of knowledge and apply that to any particular problem. The whole system and its' functionality is in one place, and probably well-defined and understood. With a distributed system, each individual component can be much simpler, where the nature of this allows for easier cooperation between those entities. Unfortunately however, communications between large numbers of simpler entities can become just as complicated as a single system performing more complex operations on its own. Because a complex problem is naturally broken down into simpler ones as part of the problem-solving process, a distributed system is not that different to a centralised one when solving the same problem. The distributed system is possibly modelled in a more modular way, which allows each component to behave in a more independent way. This is particularly useful if the operation of the system is not fully understood. In that case, the basic elements or entities of the problem can be modelled individually and then allowed to interact with each other, in the hope that they can find a suitable solution. The distributed system is also by nature more stochastic and will therefore be able to perform actions that are not predictable but are based on the current dynamic state of the system. It is more flexible in that respect.

Systems can also use search processes that evaluate incomplete or partial information. The expectation is to find a better solution, by obtaining a more complete picture through many smaller but related evaluations. Computer chess, for example, uses search processes to evaluate single positions based on imperfect evaluations. Because so many positions are evaluated however, it is able to build up a relatively reliable picture of the whole situation through these incomplete evaluations. The computer programmer would not be able to tell the system exactly what positions to evaluate, which is down to the search process itself. So this lack of knowledge is compensated for by many more evaluations and interactions that simply reveal more information from what was present in the original problem specification. The human would be expected to 'know' what the computer 'finds' through its search, although, even this is an abstract idea. The human knows more because he/she can access other information, through a different search process. Therefore, if looking at the whole search process as a single entity, they might not be so different after all. Search methods are ideal for a computer that can perform many calculations per second, but the whole process appears to lack something for modelling the human brain exactly. For these individual entities, the level of any real intelligence is still only at the entity level itself. The system probably needs some sense of itself as a whole to have intelligence at that level as well. This is the point of any stimulus feedback, to create the sense of whole from the collection of firing neurons.

Learning is also essential for intelligence. If a system cannot learn, then it is probably not intelligent. As described in the context of neural networks in section 2, it needs to be able to change internal settings through feedback. Through the manipulation and use of knowledge and rules, different types of learning process have been identified. They are also recognised as being intelligent because they perform an act that has not been directly specified beforehand. For example, a system can be asked to retrieve a certain value from a database. It can access the database and retrieve the value, but this is not intelligent. With the addition of rules, the system can then derive information or knowledge that has not been specified directly. For example, if one assertion is the fact that John bought a shirt and another assertion is the fact that all shirts are red, then by deduction it is known that John bought a red shirt. This involves the easiest method of directly traversing rules or facts that are linked, to find an answer that is not stated explicitly. Slightly more complicated would then be; if one assertion is the fact that John only buys red things and another assertion is the fact that John bought a shirt, also by deduction, it is known that the shirt is red. This is slightly more complicated, because there is no rule directly stating that the shirt is red and so a reasoning process that knows how to combine knowledge is required to come to this conclusion. The conclusion is still known to be 100% true, however. The most complicated then is induction, which actually creates something new out of what is already known. For example, if we know that John has only bought red things so far and the system is asked to buy John a coloured shirt; induction would suggest that the system should buy a red shirt. Note that that these learning processes have evolved out of knowledge-based approaches.

Distributed systems also have learning capabilities but these are less transparent and often involve statistical processes updating numerical values. A neural network, for example, is sometimes described as a black box, because the weight values that it creates and uses would not be useful in any other context and would not be understood by a human.

For a distributed system to be able to properly describe itself, any patterns that are saved will eventually need to be mapped to a symbolic system at some level and then into a language or something similar, for communication. This is the 'physical symbol system hypothesis' attributed to Newell and Simon [18]. They noted that symbols lie at the root of intelligent action. A symbol recognises one particular entity as being different from another one and also assigns a 'tag' to that entity for identification purposes. The conscious reasoning process that we know about is at this symbolic level. Another important feature that the brain might have could be very fine comparison and/or measuring capabilities. It can possibly compare these entities or symbols very accurately and measure the level of difference; especially if they are aggregations of patterns. In a general sense, intelligence can require the following:

- There is a clear distinction between a system that is 'intelligent' and one that is able simply to repeat what it has been told.
- It is relatively easy for a computer to learn and memorise information, if it is presented in a formal way. The program can also traverse the information again relatively easily, to execute any rules or actions as required. So the problem is in inferring new information from what is known, or generalising what is known to create something new.
- This probably requires the system to be able to deal with uncertainty or unpredictability at some level. Or to look at this in a different way, it requires the system to be able to predict [8]. Hawkins argues that prediction, along with memory, are the core components of intelligence, where his conclusions were based on studying the biological brain.
- Prediction includes comparisons and measuring differences. This requires using deduction, inference, induction, learning and reasoning to derive new information, or come to new conclusions from what was previously known.

- Factors such as being flexible, dynamic and able to adapt are also essential, where a learning process is required to enable these.
- Memory is also very important, when we can then start to think in terms of knowledge.
- While the stimulus with feedback (statistical or experience-based) approaches can be used to build up the 'structure' to store intelligence, knowledge (knowledge-based) is still required to properly 'understand' it. It might then be correct to state that intelligence is required to properly manipulate the knowledge.
- Rules are possibly linked pieces of related knowledge that have been worked out and found to be consistent or useful.

## 4  General Mechanisms and Processes for Building a Solution

The previous section has given one description of two general approaches that can be used to define intelligence. Experience-based approaches are required for learning and are often associated with distributed and statistical processes. They would also be associated with the lower levels of intelligence here – more at the pattern level. Knowledge-based approaches are then required for understanding and are associated more with centralised and formal methods. They would also be associated with the higher levels of intelligence here – more at the symbolic level. These two approaches also help to define how we go about building an intelligent system. One approach is to give the machine known intelligence and study how it uses that. The other is to ask the machine to form any sort of intelligence by itself. The first approach is more knowledge-based and relies on existing information, rules and scripts, which define situations or scenarios that the machine must then use in an intelligent way. These control in a pre-determined way, how the system works. Because of that, the machine can be given a more complex algorithm to process the information with. The task is to measure how it can generalise that knowledge, or create new knowledge, from what is presented to it. This approach is useful and can be used today to build practical systems. Information is represented at the symbolic level and can therefore be understood by a human, but the process can only get so far. The underlying intelligent mechanisms are not fully understood as they are pre-defined and so the system can only operate at the level of the information that it has been presented with. There are however learning processes, such as already described, to either create new knowledge or infer something that is not specified directly. So new knowledge can be created, but its domain is restricted, as is the level of real understanding.

The second approach is the modelling of the brain more closely, in a highly distributed way, with more simplistic components. The system is not allowed any (or only minimal) existing knowledge and the task is to measure what sort of knowledge it can form by itself – simple or complex. The system starts with no real structure or rule-set and creates these out of the experience and learning. A neural network, for example, is closer to this approach. The result is something that creates its own intelligence, or is able to develop consistent patterns from apparently more chaotic looking information. This approach, by itself, is not quite as useful for building practical systems, but it is just as important for modelling real intelligence. If the mechanisms for enabling a system to create its own patterns can be understood, then this will help with processing at the higher symbolic levels as well. The system must have intelligence to be able to create these patterns and if it starts with close to zero existing knowledge, then it has created this intelligence for itself. If the underlying knowledge has been created internally, then the hope would be that there is a better understanding of what it is and therefore the knowledge can be used in a more flexible way.

The ideas of tacit or explicit knowledge also address this [10]. Explicit knowledge is knowledge that can be codified, or represented in a format that can be understood by a machine. This would include a formal definition or representation of the knowledge. Tacit knowledge is knowledge held in the minds of humans that cannot be easily codified and stored on a computer. This knowledge has a personal quality created more from experience and often, this sort of knowledge is key. If a computer is allowed to generate its own knowledge, then the exact nature of it might not be completely transparent, when it can be compared more closely to tacit knowledge. For example, a chair can be described to a computer as having 4 legs, a seat and a back. We can generalise this to recognise chairs with many different shapes and forms, but we would not be able to codify those differences completely for a computer. We use our tacit knowledge to recognise the different chair shapes, sometimes based on context and this is what is missing from the programmed computer.

So to summarise, before you can reason about a concept, you have to understand what the concept is and before you can understand what it is you have to be able to distinguish it from a different one. There are still problems with this first step - for a system to autonomously learn unknown patterns or concepts for itself. Knowledge-based approaches pass over this problem, by defining these already. This point was also written about by Turing, where the following example might explain the problem:

*Scenario 1*:

*Person* (shows a tree picture): this is a tree.

*Computer*: OK.

*Person*: can you describe the object that I just showed to you?

*Computer* (accesses its database): a tree is a large woody perennial plant with a distinct trunk giving rise to branches or leaves at some distance from the ground.

*Person*: (shows a different tree picture): what is this?

*Computer*: I don't know.

*Scenario 2*:

*Computer* (after looking at lots of pictures): that looks like Object A.

*Person*: can you describe the object?

*Computer* (using own knowledge): it has a long rectangular part, smaller ones extending from that and then pointy circular objects at the end of those.

*Person* (shows a different tree picture): what is this?

*Computer*: that also looks like Object A.

If building an intelligent system, some or all of the following probably need to be part of a final model:

- To derive or induce new information, the system must be autonomous. At the lowest level, it must be able to form new knowledge or concepts by itself.
- To generate understanding, it must also be able to properly link the correct knowledge or concept parts together, so that a thinking process can follow the correct path of information.
- Pattern recognition/comparisons and the accurate measuring of differences is also critical, to allow the system to tell different entities apart.

- Symbolic reasoning is also necessary, requiring different layers of abstraction.
- The role of a stimulus by itself should not be underestimated, as our emotions, feelings and therefore preferences are controlled by that. Turing's paper notes that intelligence is not just a logical calculation, but also something such as the question 'what do you think of Picasso?'
- Rules are required. This is higher-level knowledge that links individual pieces in a constructive way. A reasoning process can create rules, where favourable/unfavourable feedback can determine the links.
- Intelligent conclusions can be at an individual level or at a societal level and influenced by knowledge of other rules or responses, etc. For example, I feel good if I eat all of the ice cream, but then everybody else is angry and so I get a bad response overall. The rule – do not eat all of the ice cream. To emphasise the point that rules are not rigid entities that everybody obeys in the same way, Turing wrote about 'laws of behaviour' as well as rules that should always be obeyed.
- Therefore, feedback is also required, as part of a learning process. Turing also emphasised this, in particular, through the example of teaching a computer more like educating a child. In that example, through evolutionary learning processes, the system is eventually able to realise some level of intelligence for itself.
- Existing knowledge is also allowed, through logical propositions (Turing), for example.

At the moment, it is not practical to try to model the brain exactly, with thousands or more neurons, all firing together. It is therefore difficult to reproduce the exact conditions under which the brain works. Because of this, adding some existing knowledge or intelligence to help the system to understand is probably required, with results then measured against what the system can do with that. Scripts can be used to help. Alternatively, much more simple processes could possibly be learned at a neuronal level, simply to see how they work. The brain eventually needs to be able to reason in a symbolic way, creating more complex concepts from linking simpler ones. Memory is the place where the learned concepts are stored and time is probably also a key element, as it allows us to form the logical associations between entities more easily. Turing described this in terms of discrete-state machines. A problem he notes with the machine is the fact that it 'is' discrete.

A more fuzzy system might do better. The machine is made of exactly defined states and concepts, but the brain would require overlapping and generalisations of these. Certain entities belong to more than one thing and as a result also represent more than one thing (context). Turing argued that where one machine fails, another might succeed, so combining these into a single machine should do both equally well. He also writes about a continuous rather than a discrete machine. Note however that state machines work at the symbolic level.

## 5  Related Work

As this paper is about Turing's work specifically, a more detailed summary of the history of Artificial Intelligence does not seem appropriate. The author however will take the opportunity to note his own cognitive model. There are many texts on artificial intelligence systems and technologies. The first place to look would be a general textbook on artificial intelligence itself, for example [20]. This section describes a cognitive, or intelligent, model that the author is currently working on [3][4][5]. It was developed from trying to optimise a network for information retrieval, when it became clear that more cognitive processes were also possible. The model structure is based mostly on feedback, or 'stigmergy' [2][15] and is also highly distributed. It is therefore in the spirit of modelling a real brain, where a diagram of the model is shown in Figure 4. One key difference with this model is the fact that it can process information as patterns, but at a symbolic level. Instead of the neural network, or cognitive model, being described as a black box or in terms of statistics; the internal workings of the model can be understood by a human user through its symbolic representations. This allows for more human-like reasoning processes to be performed. Since the paper [18], this has been noted as one of the goals of AI.

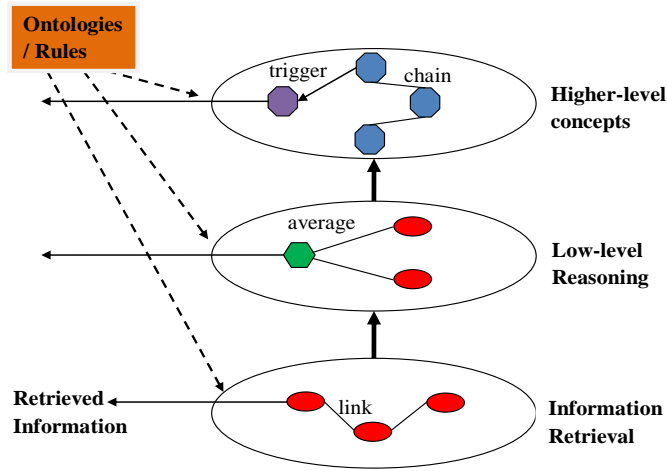

**Figure 4. Cognitive Model with three levels of complexity [5].**

This model contains three different levels of intelligence. The first or lowest level allows for basic information retrieval that is optimised through dynamic links. The linking mechanism works by linking nodes that are associated with each other through the use of the system. While it is based on the stigmergic process of linking through experience, this could also be called Hebbian [9]. Stigmergy is a very simple way of allowing components to organise themselves based on reactions to their environment. As it works through feedback, the individual components require very little intelligence or knowledge themselves. They are only required, for example, to increase the strength of a link when stimulated. A major advantage of stigmergy, or related methods, is its flexibility – the link will be reinforced in the same way, regardless of what the stimulus source is, making it generic. The knowledge or information being added to a network may not be known beforehand and so the organisation needs to be as flexible as possible. Hebb was able to study this type of behaviour in the human brain. He noticed that when an axon of cell A is near enough to excite cell B and repeatedly or persistently takes part in firing it, some growth process or metabolic change takes place in one or both cells such that A's efficiency, as one of the cells firing B, is increased. This is often paraphrased as 'Neurons that fire together wire together.' and is commonly referred to as Hebb's Law, with the linking mechanism called Hebbain. The main difference between these is the fact that stigmergy results from inputs between external and possibly independent sources, while Hebbs law results from inputs between internal and probably dependent sources. Ants for example, behaving independently of each other, can collectively perform complex tasks through stigmergic processes. The linking of the neurons in the human brain is more of an internal and related process.

The second level in the model performs simplistic aggregation or averaging operations over linked nodes. The idea being that nodes have been linked through intelligent feedback and therefore averaging over those links should be better than averaging over every random answer. The third level is more cognitive. It tries to realise more complex concepts autonomously, or independently, by linking together associated individual concepts. These links form new and distinct entities, and are not just for optimisation purposes. It also attempts to then link the more complex entities, so that a form of thinking can occur. One cluster of linked nodes, when realised, might trigger another cluster and so on. As this flows through different concepts, the network begins to realise things for itself and performs a certain level of thinking.

The first level has been tested extensively and shown to perform a good optimisation of the network. Test results ranged from a 30% reduction in search with almost no loss in the quality of answer, to possibly 80-90% reduction in the search, with maybe 5-10% loss in the quality of answer. The second level has been tested less but also shown to work. As would be expected, averaging over the linked sources only should produce a better total than averaging over all possible answers; because the nodes are linked through a process that tries to maximize the link value. The third level is the current area of research interest and some promising results have been achieved [3]. The problem is that a certain amount of randomness must be accommodated for, where the system would not be given the information exactly, but needs to perform some level of guess work. Statistical processes allow it to filter out the less likely connections and to keep the more likely ones. Two new-looking clustering algorithms [3] have been developed. These are important because they can be used as part of an autonomous system and they can also allow for a certain amount of noisy input - 10-20% already. They are also very lightweight and so are suitable for unsupervised online processing. It is these clustering processes that have led to the conclusion that a neural network architecture should be the direction for further research.

While not the original intention, this model does map loosely onto the structures that have been described. The middle layer can produce aggregated values that might be compared to the stimuli produced from aggregated patterns. The top layer can then receive or recognise different configurations of these and process them accordingly, similar to what the neocortex would do. So while only individual concepts and clusters of individual concepts have been considered so far, groups of aggregations might also be considered. The arrows between the levels represent a direction of increasing intelligence. It is likely that communication between these levels would flow in both directions. The idea of a trigger has not been worked out fully yet. It is probably related to a memory component and also a set of values or conditions under which one concept group would trigger or activate another one. In this sense, the path description associated with the linking process could be relevant. A path of concept types with related values can be associated with any link between two nodes.

The option of presenting scripts to the system has also been looked at. This is relatively easy to do and the system can learn the script and therefore know what a rule or a trigger should be. It is then only a matter of traversing this knowledge again to activate a trigger. So the problem would be to try and determine if the system can create its own rules or triggers that are not part of the original script, or if it can create the script triggers when some of the information is missing. The figure also shows an ontology or rule-base that can be used to present existing knowledge to the system. This is valid, because we also receive certain information in that form and are not expected to realise everything empirically. So research into the top, more intelligent, level has only started, but the results appear promising. One or two new-looking discoveries have been made that should help to overcome certain stumbling blocks of the past. Other work related to these ideas could include [1], [6] or [17], for example.

## 6   Conclusions

These conclusions include some of the author's own opinions, based on his limited knowledge of the real brain, but consistent with what has already been written. An attractive feature of assigning such importance to state changes, or stimulus changes, is that the individual neurons do not then require real intelligence themselves, or at least, the intelligence mechanism is now understood to be the state change that we can better understand. So the intelligence is linked to the collective chemical reactions that occur and also possibly to the very nature of a human. State changes would excite cells, which could drive our quest for new knowledge. If our state is changed in a favourable way, it makes us feel better. The brain might feel this sort of thing, even on its own. Fortunately, these reactions can also be made autonomously and so we do not have to rely completely on our environment. Then internally, the memory or some other brain area, knows the favourable/unfavourable reactions and tries to re-create them again, probably resulting in further feedback to itself. If different patterns then get linked through these reactions, even if this has not been a part of reality, the memory can still store the result to be used again. I like to think about 'A', but you like to think about 'B', for example.

The ability of the brain to make accurate comparisons is also critical, as has been written about before ([8], for example). It might be important for realising mathematical or complex operations through the feedback of results. This is probably how maths started, with somebody noticing that two piles of stones were twice as large as one pile of stones. For example, a human has worked out that two times one (stone) is twice the size of a single one (stone). The brain understands what 'one' is, at some symbolic level, and can easily associate and compare two of these symbols. This would then need to be abstracted for larger calculations, once the general understanding had been learnt. The author has also wondered why something such a driving a car is a skill that almost anybody can do, when you consider the accuracy level that is required. Almost without thinking, we do not crash into the car in-front, but measure and control our distances very easily.

So a very general rule is learned and then applied in many different scenarios. Possibly, objects from memory can be retrieved and applied to a learned rule, with feedback determining the result (see also [6], for example). Compare this to nouns and verbs in our language. Positive or recognised feedback would reinforce some association, while negative or no feedback would not register a result. An explanation of how a brain-like system can learn mathematics mainly through a stimulus process would go a long way to allowing us to model the real brain. The question might be – how much does a 'eureka' moment play in our ability to work things out. The following is also interesting for suggesting a largely mechanical process for the main brain engine: If the main brain body is purely mechanical, then it might even fire when damaged, without any consideration for the person, resulting in a painful stimulus when the damaged part is entered or interpreted. If damaged areas do fire and are not shut down or avoided, then does this suggest an unintelligent process? Why would the brain intentionally hurt itself, unless it did not know that it was doing so? Some sort of controlled process must be involved in the brain construction however, which suggests some level of controlling intelligence. The problem is really how the brain structure is created from this mysterious and controlling process. For a mechanical answer, the stimulus again offers a solution. The brain links are mechanically stimulated to grow or link in a certain manner, through the feedback that is most strongly felt.

Turing noted a lot of the problems that are still relevant today. Modelling as a state machine looks appropriate as it may be internal state changes that allow us to tell differences, resulting in intelligence. A time element is also associated with state machines. The author would suggest however that starting with a state machine is not best, but rather, the final product would be more like one. The declaration that if the problem can be described in terms of an algorithm, then it can be run on a computer, is also true. This means that if we ever figure out in a technical sense how intelligence works, it is likely that it will be transferred to a machine at a later date. Turing noted the skin-of-an-onion scenario, with layers of intelligence. The formation of patterns and then the refactoring of these into new and probably more singular ones, is essential for the formation of a symbolic level and reasoning. He also notes the importance of the other senses. While this is obvious, they are the key sources of our initial stimuli and therefore essential in the creation of our intelligence. The idea of trying to make people more intelligent through external false stimuli however, will hopefully be consigned to the waste bin.

Turing also noted that it is not practical to teach a machine in a way that the human knows and understands every step of the internal learning process. If considering state changes, the machine will make changes that the human would not know about or be able to predict. This is consistent with a learning, and therefore evolutionary process, but it means that the process must give a certain level of autonomy to the machine itself and cannot be controlled completely. The statement that a machine can only do what we tell it to is still largely true. Processes can be changed through evolution and learning, but the overall domain of their influence remains what the machine has been told. Turing argued to inject an idea into what is already known, to disturb it and allow it to ripple through the existing knowledge, in the hope of influencing or changing something. He also argued for a random element. Instead of a computer always following its instructions exactly; as part of the learning process, why not allow it to perform non-standard random acts from time to time, just so that it can receive different feedback to learn from? The problem then moves into the area of complex adaptive systems [11], with stochastic or random elements and the human teacher will definitely not be able to control that process completely.

So Turing's 'Computing Machinery and Intelligence' paper is still relevant and important today. While he wrote in a general sense, research since has been able to define the problem much more formally, but the basic premises are still the same. There have been successes in one area or another, but a comprehensive solution for intelligence has not yet been realised. It might be incorrect however to think that just because a machine is mechanical, it can never realise true intelligence. One other question would be - just how mechanical are our own brains? Turing also wrote about the theological argument against machines ever realising true intelligence, but was strongly against it. The idea that our intelligence could be based largely on stimuli is probably not an attractive one in that respect. Religious beliefs, for example, suggest that we should stay away from certain stimuli, but internal ones would possibly be OK. It is also the case that a machine cannot feel in the same way as a human and therefore, it would be difficult to model this sort of process properly in a machine. Only a 'living' organism could then have intelligence. This would be a key stumbling block to modelling intelligence properly - the feedback and evaluation mechanisms are still not real enough, but the correct algorithm simply needs to be found.

## References

[1] Fu, L. (1999). Knowledge Discovery based on Neural Networks, Communications of the ACM, Vol. 42, No. 11, pp. 47 – 50.
[2] Grassé P.P. (1959). La reconstruction dun id et les coordinations internidividuelles chez Bellicositermes natalensis et Cubitermes sp., La théorie de la stigmergie: essais d'interprétation du comportment des termites constructeurs, Insectes Sociaux, Vol. 6, pp. 41-84.
[3] Greer, K. (2011). Symbolic Neural Networks for Clustering Higher-Level Concepts, NAUN International Journal of Computers, Issue 3, Vol. 5, pp. 378 – 386, extended version of the WSEAS/EUROPMENT International Conference on Computers and Computing (ICCC'11).
[4] Greer, K, (2009). A Cognitive Model for Learning and Reasoning over Arbitrary Concepts, The 2nd International Symposium on Knowledge Acquisition and Modeling (KAM 2009), Nov 30 – Dec 1, Wuhan, China, 2009, pp. 253 - 256.

[5] Greer, K. Thinking Networks – the Large and Small of it: Autonomic and Reasoning Processes for Information Networks, LuLu.com and online at books.google.com, 2008, ISBN: 1440433275.

[6] Grossberg, S., Carpenter, G.A. and Ersoy, B. (2005). Brain Categorization: Learning, Attention, and Consciousness, Proceedings of Intemational Joint Conference on Neural Networks, Montreal, Canada, July 31 - August 4, pp. 1609 - 1614.

[7] Gurney, K. An Introduction To Neural Networks, Taylor and Francis and online at books.google.com, 1997.

[8] Hawkins, J. and Blakeslee, S. On Intelligence. Times Books, 2004.

[9] Hebb, D.O. The Organisation of Behaviour, 1994.

[10] Hildreth, P.J. and Kimble, C. (2002). The duality of knowledge, Information Research, 8(1), paper no. 142   [Available at http://InformationR.net/ir/8-1/paper142.html].

[11] Holland, J. Hidden Order: How Adaptation Builds Complexity. Reading, MA: Perseus, 1995.

[12] IBM. (2003). An Architectural Blueprint for Autonomic Computing, IBM and Autonomic Computing.

[13] Jennings, N.R. (2000). On agent-based software engineering, Artificial Intelligence, Vol. 117, pp. 277 – 296.

[14] Kohonen, T. (1990). The self-organizing map, Proceedings of the IEEE, Vol. 78, Issue 9, pp. 1464 - 1480, ISSN: 0018-9219.

[15] Mano, J.-P., Bourjot, C., Lopardo, G. and Glize, P. (2006). Bio-inspired Mechanisms for Artificial Self-organised Systems, Informatica, Vol. 30, pp. 55 – 62.

[16] McCulloch, W.S. and Pitts, W. (1943). A logical calculus of the ideas immanent in nervous activity, Bulletin of Mathematical Biophysics, Vol. 5, pp. 115 - 133.

[17] Minsky, M. (1990). Logical vs. Analogical, or Symbolic vs. Connectionist, or   Neat vs. Scruffy. In Artificial Intelligence at MIT, Expanding Frontiers, Patrick H. Winston (Ed.), Vol.1, MIT Press. Reprinted in AI Magazine, Summer 1991.

[18] Newell, A. and Simon, H.A. (1976). "Computer science as empirical inquiry: symbols and search", Communications of the ACM, Vol. 19, No. 3, pp. 113 – 126.

[19] Nikolopoulos, C. Expert Systems: Introduction to First and Second Generation and Hybrid Knowledge Based Systems, Marcel Dekker, Inc. New York, NY, USA, 1997, ISBN:0824799275.

[20] Rich, E. and Knight, K., Artificial Intelligence, McGraw Hill, Inc. 1991.

[21] Rojas, R. Neural Networks: A Systematic Introduction. Springer-Verlag, Berlin and online at books.google.com, 1996.

[22] Turing, A. (1950), "Computing Machinery and Intelligence", Mind LIX (236): 433–460, ISSN 0026-4423.

[23] Vogels, T.P., Kanaka Rajan, K. and Abbott, L.F. (2005). Neural Network Dynamics, Annu. Rev. Neurosci., Vol. 28, pp. 357 - 376.

[24] Weisbuch, G. (1999). "The Complex Adaptive Systems Approach to Biology", Evolution and Cognition, Vol. 5, No. 1, pp. 1 - 11.

[25]    Widrow, B. and Lehr, M. (1990). 30 Years of adaptive neural networks: perceptron, Madaline and backpropagation, Proc IEEE, Vol. 78, No. 9, pp. 1415-1442.